\pdfoutput=1
\documentclass{llncs}
\usepackage[utf8]{inputenc}
\usepackage{amssymb}
\usepackage{amsmath}
\usepackage{graphicx}
\graphicspath{{Graphics/}}
\usepackage{dsfont}
\usepackage{color}

\newcommand{\threepartdef}[5]
{
  \left\{
    \begin{array}{lll}
      #1 & \mbox{if } #2 \\
      #3 & \mbox{if } #4 \\
      #5 & \mbox{otherwise }
    \end{array}
  \right.
}

\definecolor{myred}{rgb}{.8,.0,.0}

\begin{document}

\title{Feature learning based on visual similarity triplets in medical image analysis: A case study of emphysema in chest CT scans}
\author{Silas Nyboe Ørting\inst{1}
  \and Jens Petersen\inst{1}
  \and Veronika Cheplygina\inst{2} 
  \and Laura H. Thomsen\inst{3}
  \and Mathilde M W Wille\inst{4} 
  \and Marleen de Bruijne\inst{1,5}}
\institute{Department of Computer Science, University of Copenhagen, Copenhagen, Denmark, \email{silas@di.ku.dk}
	\and Medical Image Analysis (IMAG/e), Department of Biomedical Engineering, Eindhoven University of Technology, Eindhoven, The Netherlands
    \and Department of Internal Medicine, Hvidovre Hospital, Copenhagen Denmark
	\and Department of Diagnostic Imaging, Bispebjerg Hospital, Copenhagen, Denmark
	\and Biomedical Imaging Group Rotterdam, Departments of Radiology and Medical Informatics, Erasmus MC - University Medical Center Rotterdam, The Netherlands}

\maketitle              

\begin{abstract}
Supervised feature learning using convolutional neural networks (CNNs) can provide concise and disease relevant representations of medical images. However, training CNNs requires annotated image data. Annotating medical images  can be a time-consuming task and even expert annotations are subject to substantial inter- and intra-rater variability. Assessing visual similarity of images instead of indicating specific pathologies or estimating disease severity could allow non-experts to participate, help uncover new patterns, and possibly reduce rater variability. 
We consider the task of assessing emphysema extent in chest CT scans. 
We derive visual similarity triplets from visually assessed emphysema extent and learn a low dimensional embedding using CNNs. We evaluate the networks on 973 images, and show that the CNNs can learn disease relevant feature representations from derived similarity triplets.
To our knowledge this is the first medical image application where  similarity triplets has been used to learn a feature representation that can be used for embedding unseen test images.
\end{abstract}
\section{Introduction}
Recent years have demonstrated the enormous potential of applying convolutional neural networks (CNNs) for medical image analysis. One of the big challenges when training CNNs is the need for annotated image data. Annotating medical images can be a time-consuming and difficult task requiring a high level of expertise. A common issue with annotations is substantial inter- and intra-rater variability. There are many sources of rater variability in annotations, for example, level of expertise, time-constraints and task definition. A common approach to defining annotation tasks is to ask raters for an absolute judgment, ``segment the tumor'', ``count number of nodules'', ``assess extent of emphysema''.  Evidence from social psychology suggests humans in some cases are better at making comparative ratings than absolute ratings \cite{wagner1997differences,goffin2011all,jones2015peer}. Redefining annotation tasks in terms of relative comparisons could improve rater agreement.

An annotation task that is especially prone to rater variations and may be better suited for comparative ratings is visual assessment of emphysema extent in chest CT scans. Emphysema is a pathology in chronic obstructive pulmonary disease (COPD), a leading cause of death worldwide \cite{GOLD}. Emphysema is characterized by destruction of lung tissue and entrapment of air. The appearance of emphysema in CT scans can be quite varied and in many cases it is difficult to precisely define where healthy tissue starts and emphysema stops. Current visual scoring systems for assessing emphysema extent are coarse yet still subject to considerable inter-rater variability \cite{barr2012a,wille2014emphysema}. Emphysema assessment based on visual similarity of lung tissue could improve rater agreement while also improving the granularity of ratings and because it is not limited by current radiological definitions, it could be used to uncover new patterns.

Current practice for visual assessment of emphysema is to consider the full lung volume and decide how much is affected by emphysema \cite{barr2012a,wille2014emphysema}. Comparing visual similarity of several 3D lung volumes simultaneously could be a difficult and time-consuming task, leading to worse rater agreement compared to assessing each volume by itself. Comparing visual similarity of 2D slices is a much easier task that could even be performed by non-experts with a little instruction. Simplifying the task to this degree opens the possibility of substituting medical experts with crowdworkers, leading to dramatic reductions in time consumption and costs. Crowdsourced image similarities have successfully been used for fine-grained bird classification \cite{wah2014similarity}, clustering of food images \cite{wilber2014cost} and more recently as a possibility for assessment of emphysema patterns \cite{orting2017crowdsourced}. 

There is a growing body of recent work on learning from similarities derived from absolute labels \cite{wang2014learning,schroff2015facenet} illustrating that learning from similarities can be better than learning directly from labels. The triplet learning setting used in these works is for learning from visual image similarity where ratings for a triplet of images $(x_i,x_j,x_k)$ are available in the form of ``$x_i$ is more similar to $x_j$ than to $x_k$''.  

In this work we also consider similarity triplets derived from absolute labels in the form of expert assessment of emphysema extent. However, our focus is on investigating the feasibility of learning in this setting, with the future goal of learning from actual visual similarity assessment of lung images.  We aim to learn descriptive image features, relevant for emphysema severity assessment, directly on the basis of visual similarity triplets. We investigate if CNNs can extract enough relevant information from a single CT slice to learn a disease relevant representation from similarity triplets. In our previous work on crowdsourcing emphysema similarity triplets \cite{orting2017crowdsourced} we did not learn a feature representation that could be used for unseen images. We believe this work is the first medical image application where similarity triplets has been used to learn a feature representation for embedding unseen images.

\section{Materials \& Method}
In this section we define the triplet learning problem and present a CNN based approach for learning  a mapping from input images to a low dimensional representation that reflects the characteristics of the visual similarity measurements.

\subsection{The triplet learning problem}
Let $\mathbf{X}$ be an image space and $x_i \in \mathbf{X}$ an image. We define a similarity triplet as an ordered triplet of images $(x_i,x_j,x_k)$ such that the ordering satisfies the triplet constraint, given by
\begin{equation}
  \label{eq:triplet-constraint}
  \delta(x_i,x_j) \le \delta(x_i, x_k)
\end{equation}
where $\delta$ is some, potentially unknown, measure of dissimilarity. Let $\mathbf{T} \subseteq \mathbf{X}^3 $ be a set of ordered triplets that satisfies (\ref{eq:triplet-constraint}). We want to find a mapping from image space to a low dimensional embedding space, $h^* : \mathbf{X} \to \mathbb{R}^d$, that minimizes the expected number of violated triplets
\begin{equation}
  \label{eq:violations}
  h^*  = \arg\min_h \mathbb{E}_{(i,j,k) \in \mathbf{T}} \left[ \mathds{1}\{ \tilde{\delta}(h(x_i),h(x_j)) \le \tilde{\delta}(h(x_i), h(x_k)) \} \right].
\end{equation}
where $\mathds{1}$ is the indicator function and $\tilde{\delta} : \mathbb{R}^d \to \mathbb{R}$ is a known dissimilarity. 

\subsection{Learning a mapping}
End-to-end learning using CNNs is a convenient and powerful method for learning concise representations of images. Optimization of CNNs is based on gradient descent and we cannot optimize (\ref{eq:violations}) directly, because the subgradient is not defined. A commonly used approach is to define a loss function based on how much a triplet is satisfied or violated
\begin{equation}
  \label{eq:loss1}
  L((x_i,x_j,x_k)) = \max\{0, \tilde{\delta}(h(x_i),h(x_j)) -  \tilde{\delta}(h(x_i), h(x_k) + C) \}
\end{equation}
where $C$ is a fixed offset used to avoid trivial solutions and encourage over-satisfying triplet constraints. Large violations can dominate the loss (\ref{eq:loss1}) and force the optimization to focus on outliers. Since we expect some inconsistencies in the similarity triplets, we consider a variant of (\ref{eq:loss1}) that bounds the loss on both sides
\begin{equation}
  \label{eq:violations-clipped-continuous}
  L((x_i,x_j,x_k)) = \mathrm{clip}_{l,u}( \tilde{\delta}(h(x_i),h(x_j)) - \tilde{\delta}(h(x_i), h(x_k)) )
\end{equation}
where 
\begin{equation}
  \label{eq:clip}
  \mathrm{clip}_{l,u}(x) = \threepartdef{0}{ x < l}{1}{x > u}{\frac{x-l}{u-l}}
\end{equation}

\paragraph{}
We consider two CNN architecture setups loosely based on VGGnet~\cite{simonyan2014very}, one with increasing and one with a fixed number of filters in each layer. In both cases a layer is comprised of zeropadding, 3x3 convolution and maxpooling. After the final layer we add a global average pooling layer, and $d$ fully connected units to obtain a $d$-dimensional embedding of the input. 
We use squared Euclidean distance as dissimilarity, i.e  $\tilde{\delta} = ||\cdot||^2_2$.

\subsection{Data}
\begin{figure}
  \centering
  \includegraphics[width=0.32\textwidth]{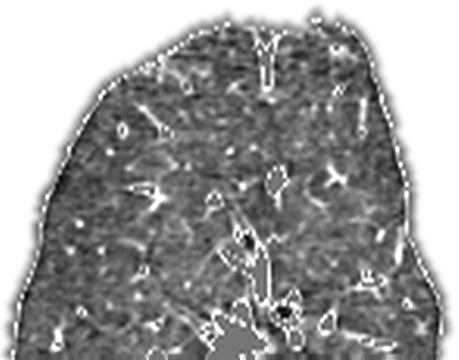}%
  \includegraphics[width=0.32\textwidth]{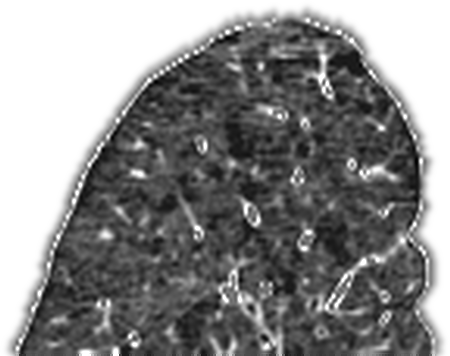}%
  \includegraphics[width=0.32\textwidth]{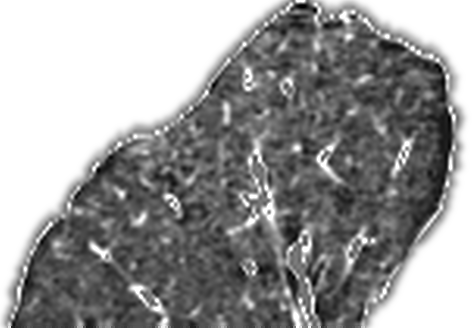}
  \includegraphics[width=0.32\textwidth]{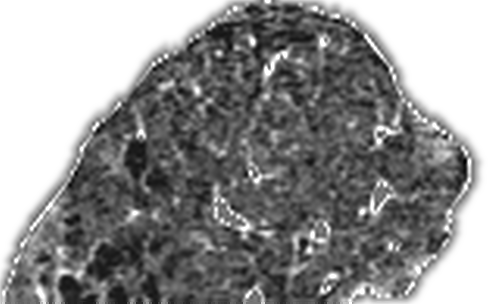}%
  \includegraphics[width=0.32\textwidth]{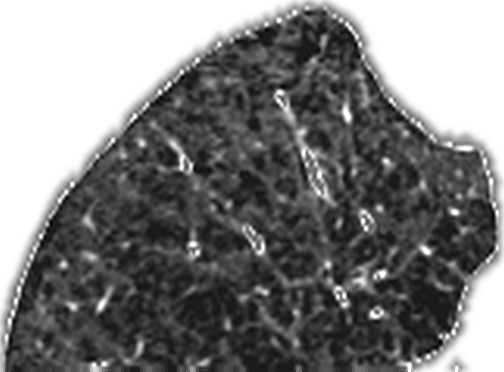}%
  \includegraphics[width=0.32\textwidth]{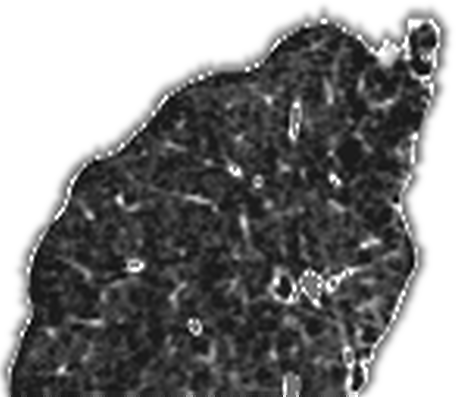}
  \caption{Example slices. From top left, visually assessed emphysema extent is 0\%, 1-5\%, 6-25\%, 26-50\%, 51-75\% and 76-100\%. Window level -780HU, window width 560HU.}
  \label{fig:example-slices}
\end{figure}

We use CT scans of 1947 subjects from a national lung cancer screening study \cite{pedersen2009dlcst} with visual assessment of emphysema extent \cite{wille2014emphysema} and segmented lung masks. Emphysema is assessed on a six-point extent scale for six regions of the lung: the upper, middle and lower regions of the left and right lung. Here we restrict our attention to the upper right region, defined as the part of the right lung lying above the carina. The six-point extent scale is defined by the intervals \{0, 1-5\%, 6-25\%, 26-50\%,51-75\%,76-100\%\}. Distribution of emphysema scores is skewed towards 0\% with about 73\% having 0\% and only about 13\% having more than 1-5\%. Example slices with varying emphysema extent are shown in Figure~\ref{fig:example-slices}.

\section{Experiments \& Results}
We split subjects randomly into a training group of 974 subjects and a test group of 973 subjects. For each experiment we then split the training group randomly in half and use one half for training and the other half for validation. Each experiment was run 10 times and we report median statistics calculated over these 10 runs. We use the same clip function for all experiments, with $[l,u] = [-0.01, 0.1]$.

\subsection{Preprocessing}
\label{sec:preprocessing}
A single coronal slice was extracted from the center of the upper right region. Bounding boxes were calculated for the lung mask of each extracted slice in the training data and all images were cropped to the size of the intersection of these bounding boxes (57 $\times$ 125 pixels). Pixels outside the lung mask were set to -800HU to match healthy lung tissue. This aggressive cropping was introduced to avoid background pixels dominating the input data. Finally all pixel intensities were scaled by $\frac{1}{1000}$ resulting in values roughly in the range $[-1,0]$.

\subsection{Selecting training triplets}
For 974 images there are close to $10^6$ possible triplets. Many of these triplets will contain very little information, and choosing the right strategy for selecting which triplets to learn from could result in faster convergence and reduce the required number of triplets needed. When class labels are available they can be used to select triplets as suggested in \cite{schroff2015facenet}. However, we are primarily interested in the setting where we do not have class labels. To understand the importance of triplet selection we compare uniform sampling of all possible triplets to sampling based on emphysema extent labels.

When selecting triplets based on emphysema extent we pick the first image uniformly at random from all images, the second uniformly at random from all images with the same emphysema extent as the first, and the third from from all images with different emphysema extent. For the third image we sample images with probability proportional to the absolute difference between the labels.

\subsection{Simulating similarity assessment}
\label{sec:simulating-similarity}
We use visually assessed emphysema extent to simulate similarity assessment of image triplets. For a triplet of images $(x_i,x_j,x_k)$ with emphysema extent labels  $(y_i,y_j,y_k)$ the ordering of the triplets satisfies 
\begin{align}
  |y_{\sigma(1)} - y_{\sigma(2)}| \le |y_{\sigma(1)} - y_{\sigma(3)}|\\
  |y_{\sigma(1)} - y_{\sigma(2)}| \le |y_{\sigma(2)} - y_{\sigma(3)}|\\
  |y_{\sigma(1)} - y_{\sigma(3)}| \le |y_{\sigma(2)} - y_{\sigma(3)}|
\end{align}
This corresponds to asking a rater to order images based on similarity.

\subsection{CNN selection}
We implemented all CNNs in Keras \cite{chollet2015keras} and used the default Adam optimizer. We searched over networks with $\{3,4,5\}$ convolution layers. We used 16 filters for the setup with a fixed number of filters, and 8,16,32,64,128 for the setup with an increasing number of filters. We used a batch size of 15 images and trained the models for 100 epochs or until 10 epochs passed without decrease in triplet violations on the validation set. We then selected the weights with the lowest triplet violations on the validation set.  We expect an untrained network with randomly initialized weights will show some degree of class separation and include it as a baseline. Table~\ref{tab:validation-performance} summarizes median validation triplet violations of the selected models and the median number of epochs used for training. Triplet selection based on emphysema results in somewhat faster convergence and slightly fewer violations compared to uniform triplet selection. The difference in median epochs between uniform triplet selection and extent based triplet selection corresponds to 7500 extra training triplets for uniform selection.
\begin{table}
  \centering
  \begin{tabular}{lccc}
    Sampling scheme & Model type & Median epochs   & Median violations\\
    \hline
    Untrained       &  F3        & --              & $46.80 \pm 0.94$\\
    Uniform         &  I4        & $23.0 \pm  7.0$ & $40.84 \pm 0.71$\\
    Extent          &  F4        & $18.0 \pm  5.0$ & $39.30 \pm 0.58$
  \end{tabular}
  \caption{Validation set performance. The letter in model type indicates {\bf F}ixed or {\bf I}ncreasing number of filters and the digit indicates number of convolution layers.}
  \label{tab:validation-performance}
\end{table}

\subsection{Triplet prediction performance}
\subsubsection{Selecting test triplets}
Because we simulate similarity assessments from class labels, the selection of test triplets will have a large influence on the interpretation of performance metrics. In our case about 71\% of subjects in the test set do not have emphysema. This implies that selecting triplets uniformly at random results in about $36\%$ of the triplets having no emphysema images. We choose to ignore these same-class triplets when measuring test performance.

In addition to the issue of same-class triplets, we are also faced with a dataset where more than $50\%$ of those subjects that have emphysema only have 1-5\% extent. Ignoring this issue will lead to performance metrics dominated by the ability of the network to distinguish subjects with very little emphysema from those without emphysema. This is a difficult task even when given access to the full volume. To more fully understand how well the network embeds images with varying levels of emphysema extent, we calculate test metrics under five different test triplet selection schemes. (1) two images with same extent and one image with different extent, (2) two images without emphysema and one with emphysema, (3) two images with 0-5\% and one with $>5\%$, (4) two images with 0-25\% and one with $>25\%$, (5) two images with $0-50\%$ and one with $>50\%$. 

Table~\ref{tab:test-performance} summarizes the results. As expected we see that the networks are much better at distinguishing between subjects with moderate to severe emphysema versus mild and no emphysema (0-5\%), than subjects with emphysema versus subjects with no emphysema (0\%). We also see that the untrained network provides decent separation of images with severe emphysema versus moderate to no emphysema (0-50\%). In all cases we see that using information about emphysema extent for generating training triplets leads to better performance compared with uniform sampling of triplets.
\begin{table}
  \centering
  \begin{tabular}{lccccc}
    & \multicolumn{5}{c}{Test triplet selection method}\\
    Sampling scheme  &  All & 0\%  & 0-5\% & 0-25\% & 0-50\% \\ 
    \hline
    Uniform          & 41.0         & 40.2         & 30.0         & 19.0         & 11.6\\
    Extent           & 39.3         & 39.0         & 26.4         & 14.6         & 9.4\\
    Untrained        & 48.5         & 48.9         & 44.3         & 37.2         & 29.2
  \end{tabular}
  \caption{Median triplet violations on test set for the selected models from Table~\ref{tab:validation-performance} using different schemes for selecting test triplets. See text for explanation of column names.}
  \label{tab:test-performance}
\end{table}

\begin{figure}
  \centering
  \includegraphics[width=0.3\textwidth,trim=65 75 35 65,clip=true]{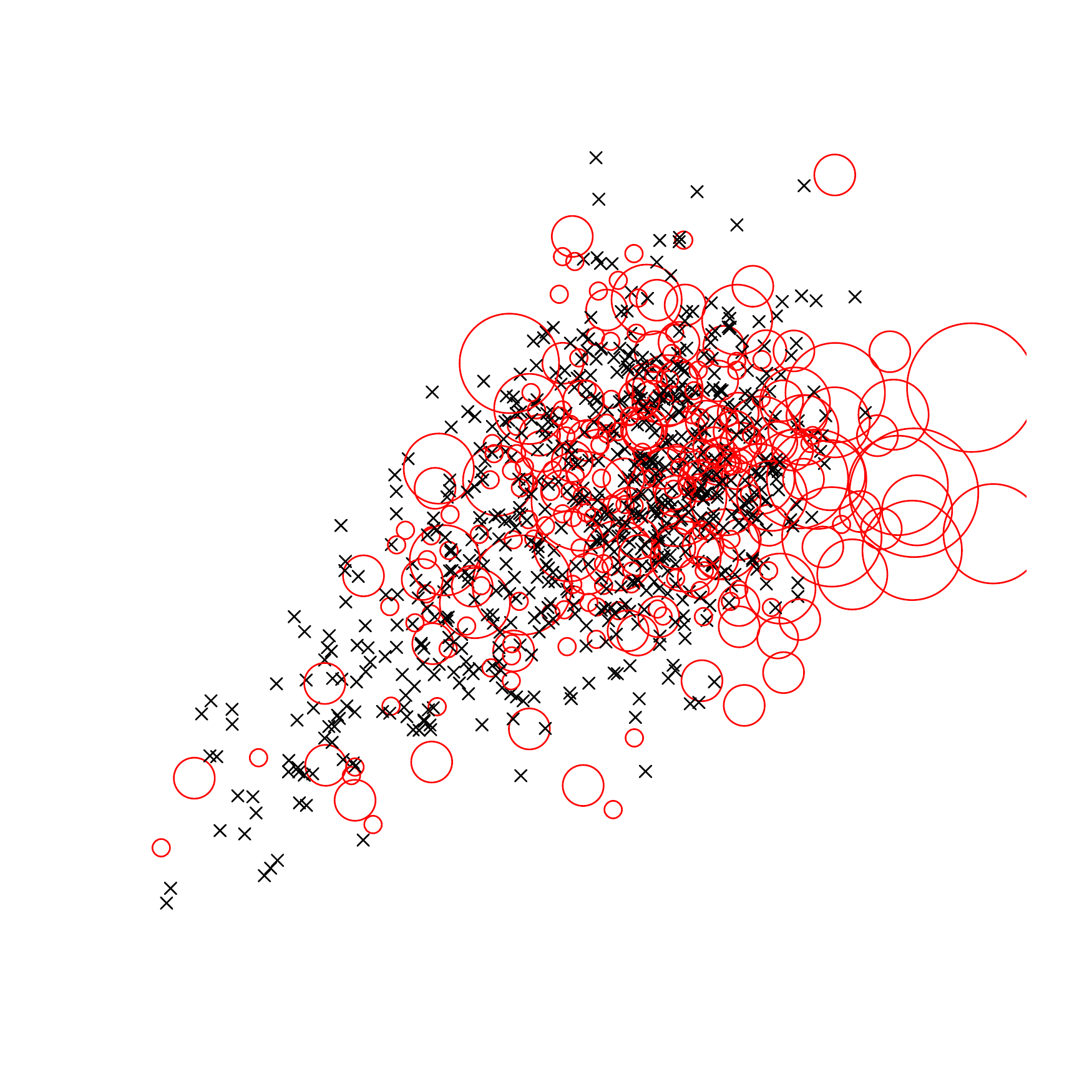}
  \includegraphics[width=0.3\textwidth,trim=65 75 35 65,clip=true]{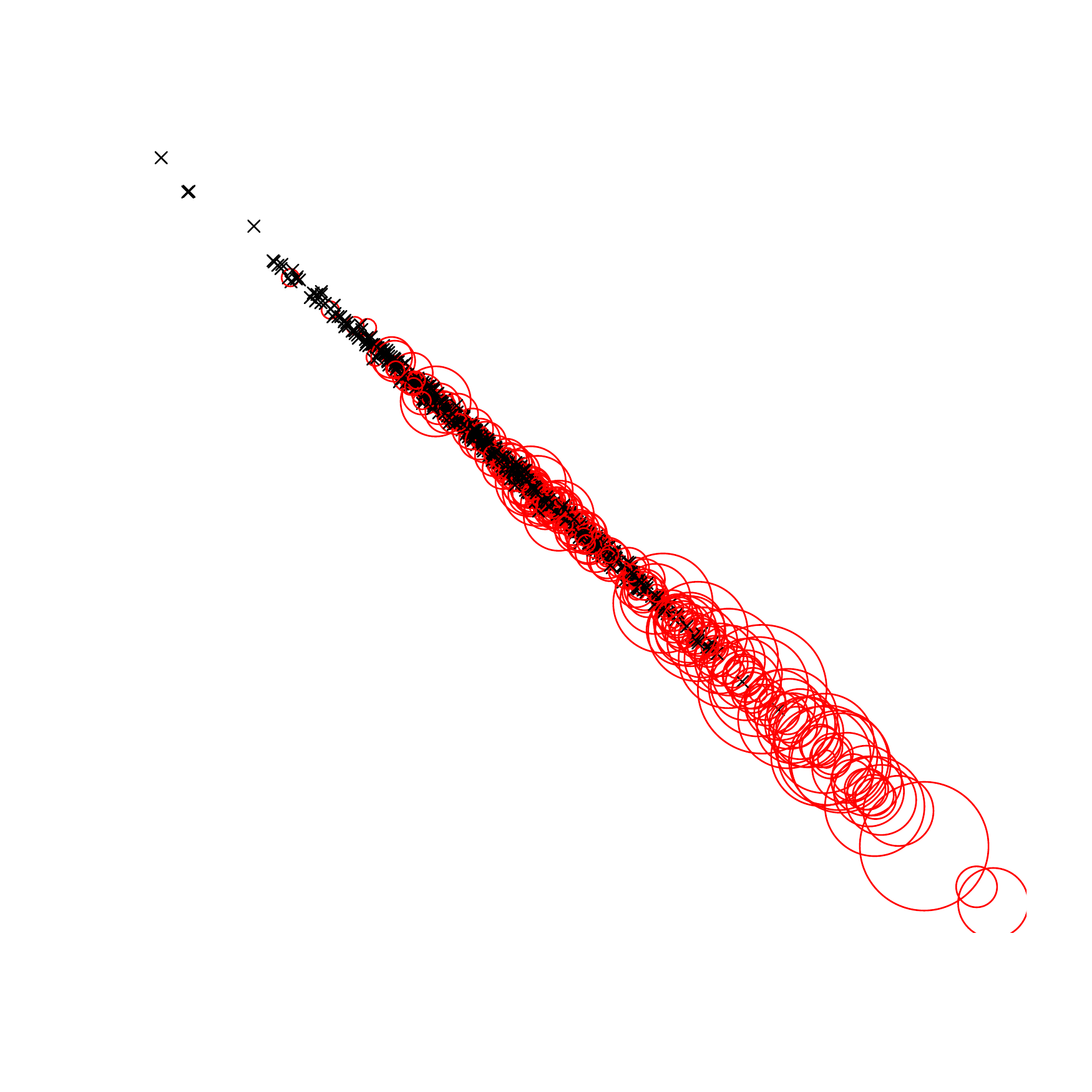}
  \includegraphics[width=0.3\textwidth,trim=65 75 35 65,clip=true]{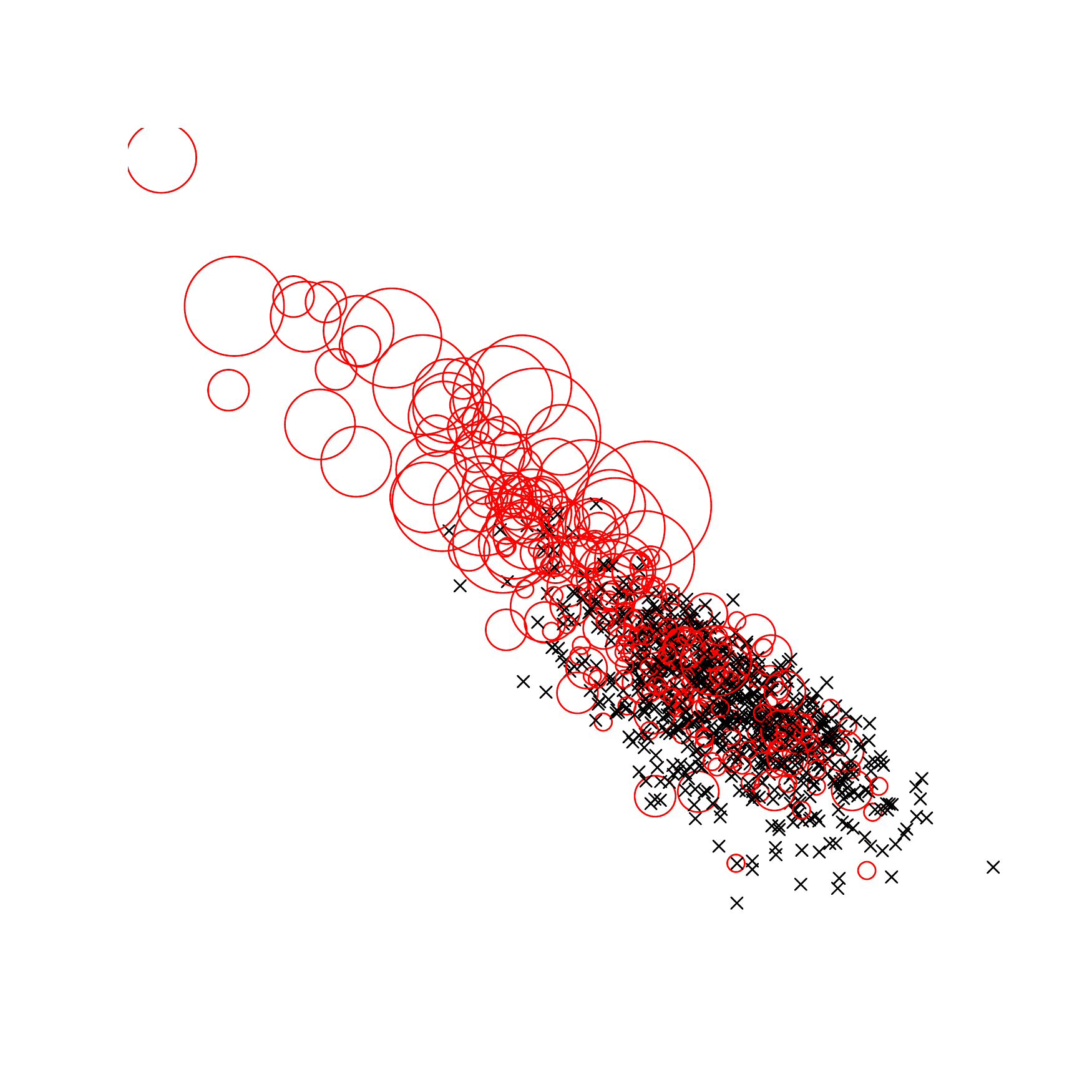}
  \caption{Example embedding of test data. Black crosses are subjects without emphysema, red circles are subjects with emphysema. Size of circle correspond to emphysema extent. From left: Untrained (48.3\% testset violations), uniform (39.5\% testset violations), visual (38.8\% testset violations).}
  \label{fig:embedding}
\end{figure}

An example embedding of the test set is shown in Figure~\ref{fig:embedding}. We used the models with best performance on the validation set to generate the embedding. Although we see significant overlap between subjects with and without emphysema, both of the trained embeddings have a reasonably pure cluster of subjects with emphysema. There is a clear tendency towards learning a one dimensional embedding. We hypothesize that several factors contribute to this tendency, (1) clipping at $[-0.01,0.1]$ encourages small distances, (2) pairwise distances for uniformly distributed points increase as the dimensionality is increased, (3) the underlying dissimilarity space, emphysema extent, is one dimensional and all triplets can in principle be satisfied by embedding unto the real line.

\section{Discussion \& Conclusion}
We formulated assessment of emphysema extent as a visual similarity task and presented an approach for learning an emphysema relevant feature representation from similarity triplets using CNNs. We derived similarity triplets from visual assessment and investigated the importance of selecting informative triplets. 

It is slightly surprising that a single cropped 2D slice contains enough information for the level of separation illustrated by the embeddings in Figure~\ref{fig:embedding}. This shows that learning can be accomplished from simple annotation tasks. However, there are likely instances where the particular slice is not representative for the image as a whole, which may explain why there is a large overlap between subjects with and without emphysema in Figure~\ref{fig:embedding}. We suspect that with triplet similarities based on individual slice comparisons, class overlap would be less.

As a proof of concept, in this work we simulated slice similarity assessment from experts' emphysema extent scores.  Potentially such triplets could be gathered online via crowdsourcing platforms such as Amazon Mechanical Turk. Our previous results \cite{orting2017crowdsourced} showed that crowdsourced triplets could be used to classify the emphysema type (rather than extent) with a better than random performance. Preliminary results indicate that the crowdsourced triplets are too few or too noisy for training the proposed CNNs. However, we expect that improving the quality and increasing the quantity of crowdsourced triplets will allow CNNs to learn an emphysema sensitive embedding without needing expert assessed emphysema extent for training.

We investigated the importance of triplet selection and found that performance improved slightly when selecting triplets based on emphysema extent, in particularly for subjects with moderate emphysema extent (columns 0-5\% and 0-25\% in Table~\ref{tab:test-performance}). While using disease class labels to select triplets is not a viable solution, for medical images we often have access to relevant clinical information that could be used to select triplets. In the context of emphysema, measures of pulmonary function are potential candidates for triplet selection. However, our preliminary results indicate that using pulmonary function measures for triplet selection is not straightforward and can harm performance compared to uniform triplet selection.

We assumed that there is a single definition of visual similarity between the slices. However, this does not have to hold in general. For emphysema it is relevant to consider both pattern and extent as measures of similarity. The idea of having multiple notions of similarity is explored in \cite{veit2017conditional}, where different subspaces of the learned embedding corresponds to different notions of similarity. Simultaneously modeling multiple notions of similarity could lead to more expressive feature representations. Additionally, it be useful when learning from crowdsourced triplets, where some raters might focus on irrelevant aspects, such as size and shape of the lung.

In conclusion, we have shown that CNNs can learn an informative representation of emphysema based on similarity triplets. We believe this to be a promising direction for learning from relative ratings, which may be more reliable and intuitive to do, and therefore could allow the collection of large data sets that CNNs benefit from. The next step is to explore embeddings resulting from directly annotated similarity triplets. We expect such embeddings to show different notions of similarity and it will be interesting to see how these notions compare to current radiological definitions.

%
%
\bibliographystyle{abbrv}
\bibliography{refs}

\begin{thebibliography}{10}

\bibitem{chollet2015keras}
F.~Chollet et~al.
\newblock Keras.
\newblock \url{https://keras.io}, 2015.

\bibitem{barr2012a}
{COPDGene CT Workshop Group: R. Graham Barr et al}.
\newblock A combined pulmonary-radiology workshop for visual evaluation of
  {COPD}: Study design, chest {CT} findings and concordance with quantitative
  evaluation.
\newblock {\em COPD: Journal of Chronic Obstructive Pulmonary Disease},
  9(2):151--159, 2012.

\bibitem{goffin2011all}
R.~D. Goffin and J.~M. Olson.
\newblock Is it all relative? comparative judgments and the possible
  improvement of self-ratings and ratings of others.
\newblock {\em Perspectives on Psychological Science}, 6(1):48--60, 2011.

\bibitem{GOLD}
{Global Strategy for the Diagnosis, Management and Prevention of {COPD}, Global
  Initiative for Chronic Obstructive Lung Disease {(GOLD)} 2017}, 2017.

\bibitem{jones2015peer}
I.~Jones and C.~Wheadon.
\newblock Peer assessment using comparative and absolute judgement.
\newblock {\em Studies in Educational Evaluation}, 47:93--101, 2015.

\bibitem{orting2017crowdsourced}
S.~N. {\O}rting, V.~Cheplygina, J.~Petersen, L.~H. Thomsen, M.~M. Wille, and
  M.~de~Bruijne.
\newblock Crowdsourced emphysema assessment.
\newblock In {\em Intravascular Imaging and Computer Assisted Stenting, and
  Large-Scale Annotation of Biomedical Data and Expert Label Synthesis}, pages
  126--135. Springer, 2017.

\bibitem{pedersen2009dlcst}
J.~H. Pedersen, H.~Ashraf, A.~Dirksen, K.~Bach, H.~Hansen, P.~Toennesen,
  H.~Thorsen, J.~Brodersen, B.~G. Skov, M.~Døssing, J.~Mortensen, K.~Richter,
  P.~Clementsen, and N.~Seersholm.
\newblock The {Danish} randomized lung cancer {CT} screening trial--overall
  design and results of the prevalence round.
\newblock {\em Journal of Thoracic Oncology}, 4(5), 2009.

\bibitem{schroff2015facenet}
F.~Schroff, D.~Kalenichenko, and J.~Philbin.
\newblock Facenet: A unified embedding for face recognition and clustering.
\newblock In {\em Proceedings of the IEEE conference on computer vision and
  pattern recognition}, pages 815--823, 2015.

\bibitem{simonyan2014very}
K.~Simonyan and A.~Zisserman.
\newblock Very deep convolutional networks for large-scale image recognition.
\newblock {\em arXiv preprint arXiv:1409.1556}, 2014.

\bibitem{veit2017conditional}
A.~Veit, S.~Belongie, and T.~Karaletsos.
\newblock Conditional similarity networks.
\newblock {\em Computer Vision and Pattern Recognition (CVPR 2017)}, 2017.

\bibitem{wagner1997differences}
S.~H. Wagner and R.~D. Goffin.
\newblock Differences in accuracy of absolute and comparative performance
  appraisal methods.
\newblock {\em Organizational Behavior and Human Decision Processes},
  70(2):95--103, 1997.

\bibitem{wah2014similarity}
C.~Wah, G.~Van~Horn, S.~Branson, S.~Maji, P.~Perona, and S.~Belongie.
\newblock Similarity comparisons for interactive fine-grained categorization.
\newblock In {\em Proceedings of the IEEE Conference on Computer Vision and
  Pattern Recognition}, pages 859--866, 2014.

\bibitem{wang2014learning}
J.~Wang, Y.~Song, T.~Leung, C.~Rosenberg, J.~Wang, J.~Philbin, B.~Chen, and
  Y.~Wu.
\newblock Learning fine-grained image similarity with deep ranking.
\newblock In {\em Proceedings of the IEEE Conference on Computer Vision and
  Pattern Recognition}, pages 1386--1393, 2014.

\bibitem{wilber2014cost}
M.~J. Wilber, I.~S. Kwak, and S.~J. Belongie.
\newblock Cost-effective hits for relative similarity comparisons.
\newblock In {\em Second AAAI Conference on Human Computation and
  Crowdsourcing}, 2014.

\bibitem{wille2014emphysema}
M.~M. Wille, L.~H. Thomsen, A.~Dirksen, J.~Petersen, J.~H. Pedersen, and S.~B.
  Shaker.
\newblock {{E}mphysema progression is visually detectable in low-dose {C}{T} in
  continuous but not in former smokers}.
\newblock {\em Eur Radiol}, 24(11):2692--2699, Nov 2014.

\end{thebibliography}

\end{document}